\documentclass[10pt,twocolumn,letterpaper]{article}

\usepackage{cvpr}              
\definecolor{cvprblue}{rgb}{0.21,0.49,0.74}

\usepackage{multirow}
\usepackage{multicol}
\usepackage{amssymb}

\usepackage{pifont}

\definecolor{emerald}{RGB}{80, 200, 120}
\definecolor{coral}{RGB}{255, 127, 80}
\definecolor{teal}{RGB}{0, 128, 128}
\definecolor{goldenrod}{RGB}{218, 165, 32}

\definecolor{darkgreen}{RGB}{0,100,0}
\definecolor{darkred}{RGB}{139,0,0}

\usepackage[pagebackref,breaklinks,colorlinks,allcolors=cvprblue]{hyperref}

\usepackage[table]{xcolor}
\usepackage{tcolorbox}

\title{Reasoning Text-to-Video Retrieval via Digital Twin Video Representations and Large Language Models}

\author{Yiqing Shen, Chenxiao Fan, Chenjia Li, Mathias Unberath\\
Department of Computer Science, Johns Hopkins University\\
{\tt\small \{yshen92, unberath\}@jhu.edu}
}

\begin{document}
\maketitle
\begin{abstract}
The goal of text-to-video retrieval is to search large databases for relevant videos based on text queries. 
Existing methods have progressed to handling explicit queries where the visual content of interest is described explicitly; however, they fail with implicit queries where identifying videos relevant to the query requires reasoning.
We introduce reasoning text-to-video retrieval, a paradigm that extends traditional retrieval to process implicit queries through reasoning while providing object-level grounding masks that identify which entities satisfy the query conditions. 
Instead of relying on vision-language models directly, we propose representing video content as digital twins \textit{i}.\textit{e}., structured scene representations that decompose salient objects through specialist vision models. This approach is beneficial because it enables large language models (LLMs) to reason directly over long-horizon video content without visual token compression. 
Specifically, our two-stage framework first performs compositional alignment between decomposed sub-queries and digital twin representations for candidate identification, then applies LLM-based reasoning with just-in-time refinement that invokes additional specialist models to address information gaps.
We construct a benchmark of 447 manually created implicit queries with 135 videos (ReasonT2VBench-135) and another more challenging version of 1000 videos (ReasonT2VBench-100).
Our method achieves 81.2\% R@1 on ReasonT2VBench-135, outperforming the strongest baseline by >50\%-points, and maintains 81.7\% R@1 on the extended configuration while establishing state-of-the-art results in three conventional benchmarks (MSR-VTT, MSVD, and VATEX).
\end{abstract}    
\section{Introduction}

\begin{figure}[!htbp]
\centering
\includegraphics[width=\linewidth]{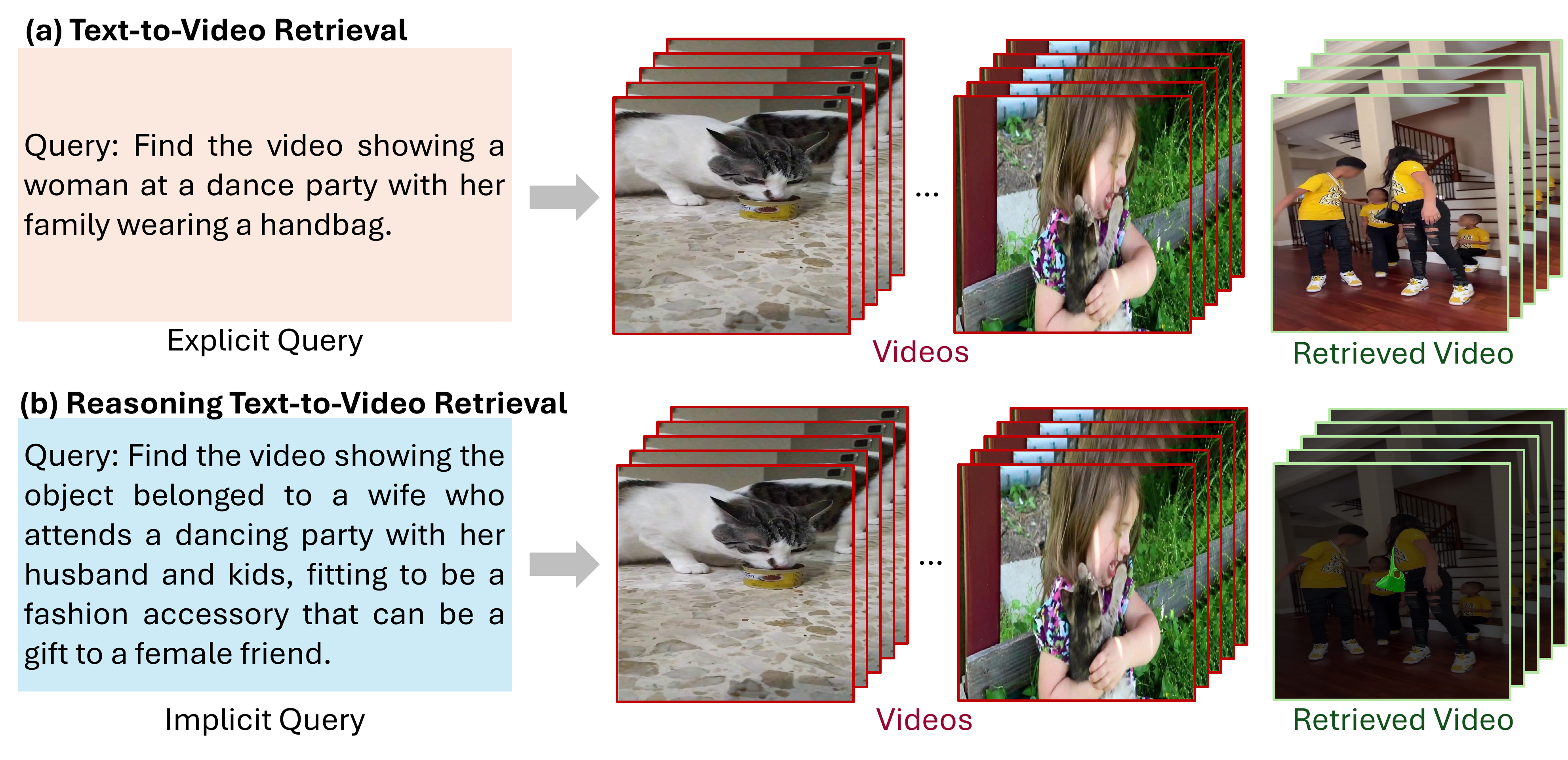}
\caption{Comparison between (a) traditional text-to-video retrieval and (b) reasoning text-to-video retrieval. 
Traditional retrieval accepts explicit queries and returns matching videos without identifying target objects. 
In contrast, reasoning retrieval interprets indirect query that demand multiple reasoning steps, while simultaneously localizing the referenced object through segmentation masks (shown in green) within the retrieved video.}
\label{fig:intro}
\end{figure}

Text-to-video retrieval (T2V retrieval) aims to retrieve targeted videos from database using natural language queries~\cite{zhu2023deep}.
It has broad applicability ranging from video search engines~\cite{annamalai2024bridging} to video recommendation systems~\cite{chaubey2025contextiq}.
Existing T2V retrieval typically operates on video-text matching, where the text queries explicitly describe visual content appearing in target videos~\cite{zhang2025text,li2023progressive}.
Specifically, the T2V retrieval method returns entire videos whose overall visual content aligns with the text query description, without identifying which specific objects within those videos satisfy the query conditions.
Moreover, this formulation cannot handle queries that encode implicit semantic, spatial, or temporal relationships that require reasoning~\cite{malik2025ravu}.
Consider retrieving videos for the query ``\textit{a cat playing with yarn}'' versus ``\textit{videos showing an animal discovering something that captures its curiosity}.''
The first query specifies concrete elements to enable straightforward matching; while the second query expresses abstract concepts that require reasoning implicitly.
To address these limitations, we introduce \textbf{reasoning text-to-video retrieval}, which extends traditional T2V retrieval in two ways, as shown in Fig.~\ref{fig:intro}: (1) supporting implicit queries that require reasoning, and (2) providing object-level grounding that identifies which specific objects within retrieved videos satisfy the query.
This formulation can better reflect real-world retrieval scenarios where users express needs through implicit descriptions and seek not only relevant videos, but also understanding of where and how the described content manifests within those videos.

However, current T2V retrieval methods cannot interpret implicit queries nor provide the object-level grounding required for reasoning T2V retrieval.
Methods such as CLIP4Clip~\cite{luo2022clip4clip} and X-CLIP~\cite{ma2022xclip} learn joint embeddings that map text and video to a shared representation space, where retrieval operates through cosine similarity matching between query and video embeddings.
When faced with implicit queries that require reasoning, these holistic embeddings fail because they cannot decompose queries into constituent reasoning steps.
Moreover, these methods produce only video-text similarity scores, which offer no mechanism to identify which objects or regions within the retrieved videos actually satisfy the queries.
Video-LLMs offer an alternative paradigm by processing video representations through large language models (LLMs) to compute relevance scores.
However, these methods compress visual information into extremely limited token sequences to fit within LLM context windows~\cite{shen2024longvu}, eliminating the spatial precision and object-level detail required for reasoning retrieval.
For example, LLaMA-VID~\cite{li2024llama} represents each frame with only two tokens, eliminating the spatial precision and object-level detail needed for accurate reasoning and grounding.
Without preserving individual object boundaries, locations, and attributes, these compressed representations cannot identify specific entities that satisfy query conditions.

To address both challenges in reasoning text-to-video retrieval, we propose a two-stage framework that converts videos into digital twin representations.
A digital twin representation is a structured scene representation that decomposes video content through various vision foundation models into explicit object-level components~\cite{jit,position}.
This transformation enables LLMs to perform reasoning over structured text rather than compressed visual tokens, preserving the fine-grained spatial information needed for object-level grounding.
For efficiency, we first employ compositional alignment that learns joint embeddings between LLM-decomposed sub-queries and digital twin components, enabling rapid identification of candidate videos without exhaustive reasoning over the entire database.
The top candidates then undergo fine-grained analysis through LLM-based reasoning that matches query conditions against the structured representation, with just-in-time refinement invoking additional specialist models when information gaps arise. 
Our method returns both a ranked list of relevant videos and object-level masks specifying which entities within those videos satisfy the query conditions.

The major contributions are three-fold.
First, we define reasoning text-to-video retrieval as a task that retrieves the relevant videos based on implicit queries while localizing objects that can fulfill the query.
Second, we introduce a two-stage reasoning text-to-video retrieval method that decomposes implicit queries into atomic sub-queries for compositional alignment with digital twin representations, followed by LLM-based reasoning with just-in-time refinement to identify and ground objects satisfying complex query conditions.
Third, we construct a reasoning text-to-video retrieval benchmark with two configurations, namely \textit{ReasonT2VBench-135} containing 135 videos with 447 implicit text queries, and \textit{ReasonT2VBench-1000} which adds 865 irreverent distractor videos to further test semantic discrimination.
All implicit text queries are accompanied by ground-truth video and instance-level ground-truth masks, which are validated through a two-stage verification pipeline.
\section{Related Works}

\begin{figure*}[!t]
\centering
\includegraphics[width=\linewidth]{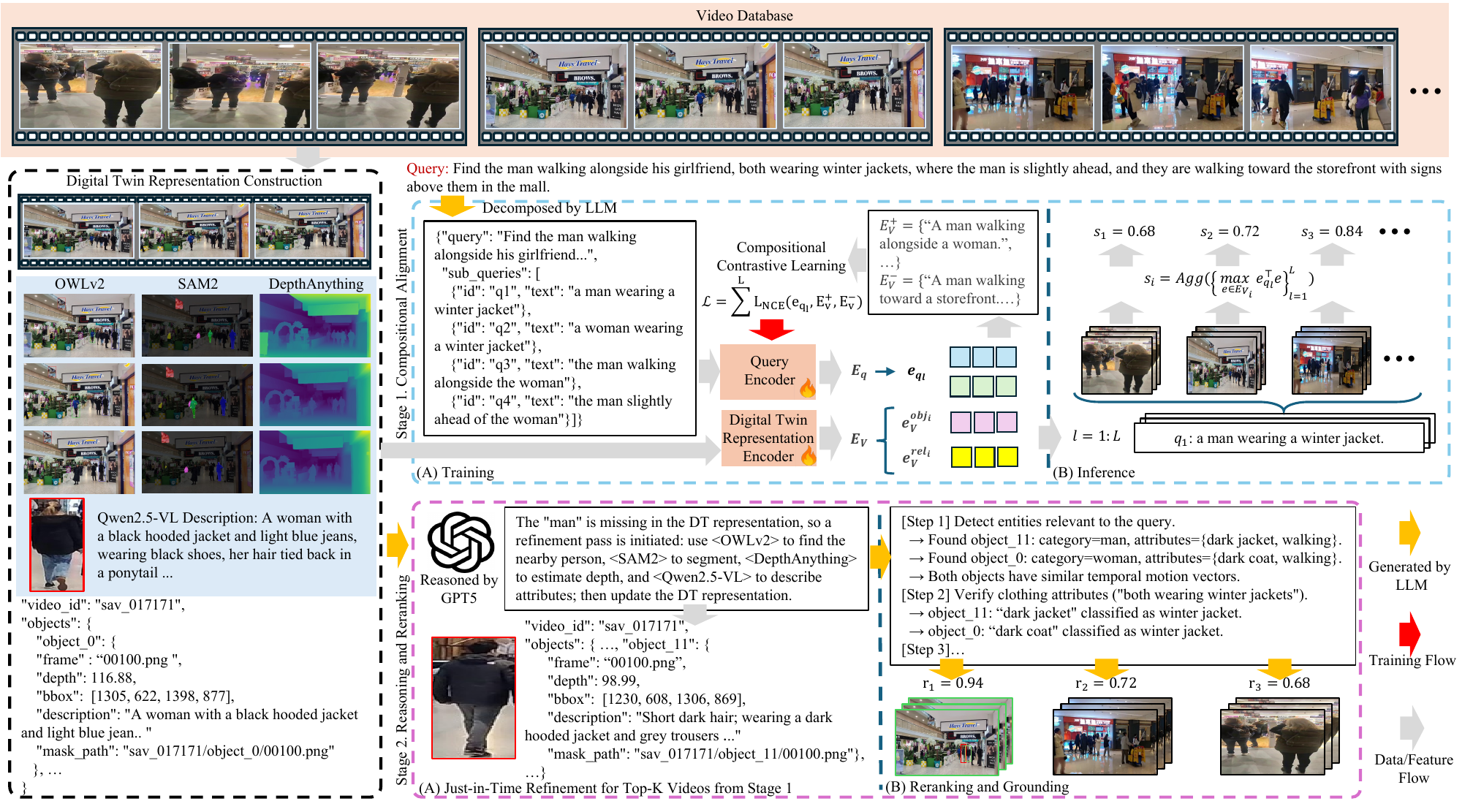}
\caption{Overall framework of the proposed reasoning text-to-video retrieval method.}
\label{fig:framework}
\end{figure*}

\paragraph{Text-to-Video Retrieval}
Initial methods for text-to-video retrieval like VSE++ employ dual encoding, where videos and text are independently encoded before being projected into common spaces for similarity computation~\cite{faghri2017vse++}.
Similarly, CLIP4Clip adapted pre-trained image-text models to video domains through frame-level feature aggregation, computing relevance via cosine similarity between query and video embeddings~\cite{luo2022clip4clip}. 
X-CLIP extended this approach with multi-grained contrastive learning across different temporal scales~\cite{ma2022xclip}.  
The multimodal transformer (MMT) demonstrated the benefits of joint encoding in multiple video modalities, including appearance, motion, audio, and speech, through the transformer~\cite{gabeur2020multi}. 
While these methods achieve strong performance on explicit queries describing visual content, they struggle with implicit queries requiring multi-hop reasoning.
Their holistic embeddings cannot decompose complex queries into constituent reasoning steps, nor can they identify which specific objects within retrieved videos satisfy query conditions.
Temporal grounding methods such as 2D-TAN and Moment-DETR have attempted to address localization within videos, but remain confined to video-level retrieval without providing object-level grounding that reasoning tasks demand~\cite{zhang20202d, lei2021detecting}.

\paragraph{Reasoning Tasks for Visual Data}
Recent work in reasoning segmentation has shown that models can identify objects through implicit queries that require reasoning~\cite{lisa,survey}. 
For example, LISA introduced a special \texttt{<SEG>} token and embedding-as-mask paradigm within multimodal LLM (MLLM), allowing these MLLMs to produce segmentation masks from implicit queries that involve world knowledge and spatial reasoning~\cite{lisa}. 
GSVA extended these capabilities by learning multiple \texttt{<SEG>} tokens for multi-target scenarios and introducing a \texttt{<REJ>} token for explicit rejection of absent targets~\cite{gsva}. 
VISA extended reasoning segmentation to videos, showing that MLLMs can perform object-level reasoning across videos~\cite{visa}. 
These approaches show a shift from explicit matching to implicit understanding in segmentation, where models now interpret implicit semantic, spatial, or temporal relationships to identify targets~\cite{survey}.
However, reasoning segmentation focuses on localizing objects within given images or videos, while reasoning retrieval must search across databases to find relevant videos and ground the specific objects that satisfy implicit queries.

\paragraph{LLMs for Video Understanding}
LLMs have been applied to video understanding with methods that compress visual information to fit context windows. 
Video-ChatGPT combined vision encoders with LLMs for video conversation~\cite{maaz2023video}, while LLaMA-VID represented each frame with only two tokens~\cite{li2024llama}. 
For long videos, R-VLM uses learnable retrieval to select relevant chunks~\cite{xu2023retrieval}, and VideoRAG builds graph-based knowledge structures for cross-video reasoning~\cite{ren2025videorag,luo2024video}. 
However, token compression sacrifices the spatial precision needed for grounding at the object level.
Digital twin representations offer an alternative by decomposing scenes into explicit object-level components~\cite{jit,position,miccai}, preserving fine-grained spatial information through specialist vision models while enabling LLMs to perform high-level reasoning.

\section{Methods}

\paragraph{Problem Formulation}
Given a video database $\mathcal{V} = \{V_1, V_2, ..., V_N\}$ and an implicit text query $q$ that requires reasoning to interpret, the reasoning text-to-video retrieval task aims to retrieve a ranked list of relevant videos along with object-level grounding masks. 
Formally, we seek a retrieval function $f: (q, \mathcal{V}) \rightarrow \{(V_i, \mathcal{M}_i, s_i)\}_{i=1}^{K}$, where $K$ denotes the number of retrieved videos, $\mathcal{M}_i = \{m_1^{(i)}, m_2^{(i)}, ..., m_{P_i}^{(i)}\}$ represents the set of binary segmentation masks for objects in video $V_i$ that satisfy the query conditions, and $s_i \in [0,1]$ is the relevance score. 
Unlike traditional text-to-video retrieval where queries explicitly describe visual content, our task handles implicit queries requiring reasoning.

\paragraph{Overview}
We propose a two-stage retrieval method specifically for the reasoning text-to-video retrieval, as shown in Fig.~\ref{fig:framework}.
Specifically, in the pre-processing phase, we first convert all videos into digital twin representations following the design of previous work~\cite{jit}, where specialist vision models extract and structure visual information into explicit object-level components.
With the digital twin representation encoded in text, we can consequently transform this retrieval problem from direct text-video matching into reasoning and retrieval over text with LLM.
For retrieval, we first decompose each implicit text query into atomic explicit sub-queries through an LLM.
We then learn joint embeddings between these explicit sub-queries and digital twin representations using two paired encoders, which allows efficient identification of candidate videos without exhaustive reasoning over the entire database during the inference.
For the retrieved candidates, we then employ an LLM-based agent that performs reranking by matching query conditions against digital twin representation.
During this reasoning process, the agent may identify information gaps between the initial digital twin representation and the query requirements.
We therefore propose a just-in-time refinement method, where the LLM generates execution plans that invoke additional specialist models to augment the digital twin representation with query-specific details.
Finally, our method produces two outputs, namely a ranked list of relevant videos and object-level grounding masks that specify which objects within each video satisfy the query conditions.
These masks are obtained from the corresponding object mask's path in the digital twin representation.

\paragraph{Digital Twin Representation Construction}
We begin by converting each video in the database $\mathcal{V} = \{V_1, V_2, ..., V_N\}$ into a digital twin representation that explicitly decomposes visual content into structured and object-level components.
Formally, for a video $V$ with $T$ frames, the digital twin representation $\mathcal{D}_V = \{D^{(1)}, D^{(2)}, ..., D^{(T)}\}$ contains a structured scene description for each frame.
Each frame representation $D^{(t)}$ consists of detected object instances, depicted as
\begin{equation}
D^{(t)} = \{(i, c_i^{(t)}, \mathbf{a}_i^{(t)}, m_i^{(t)}, \mathbf{s}_i^{(t)})\}_{i=1}^{N^{(t)}},
\end{equation}
where $N^{(t)}$ denotes the number of instances in the frame $t$.
Each instance tuple contains: an identifier $i$ to track across frames (which should be consistent with the previous frame); semantic category $c_i^{(t)}$; attribute descriptions $\mathbf{a}_i^{(t)}$ to capture visual properties such as color and \textit{etc}; mask path $m_i^{(t)}$ to store location reference for object grounding; and spatial properties $\mathbf{s}_i^{(t)} = (x, y, d, \text{size})$ to encode centroid coordinates, depth, and object size.
We construct these representations using specialist vision models~\cite{jit}.
Specifically, SAM-2~\cite{sam2} generates instance-level segmentation masks and maintains object correspondences between frames.
DepthAnything~\cite{depthanything} estimates the depth maps per pixel that we sample at object centroids.
OWLv2~\cite{owl} assigns semantic categories to each detected instance.
MLLM, \textit{i}.\textit{e}. Qwen2.5-VL~\cite{qwen3,qwenvl}, generates natural language descriptions of object attributes, including color, texture, and appearance.
This decomposition preserves fine-grained object information and spatial-temporal relationships that would be lost in holistic video embeddings.
Finally, we serialize $\mathcal{D}_V$ in JSON format, which therefore allows LLMs to process the digital twin representation directly through their language understanding capabilities, without requiring visual encoders~\cite{jit,miccai}.

\begin{table*}[t!]
\centering
\caption{Performance comparison on \textit{ReasonT2VBench-135} across ranking metrics (R@K) and average precision (AP@K). Our method consistently outperforms traditional embedding-based retrieval and video-LLM approaches across all metrics. Values represent mean $\pm$ standard deviation. \textbf{Bold} indicates best results and \underline{underline} indicates second-best results.}
\label{tab:model_comparison}
\resizebox{0.97\linewidth}{!}{%
\begin{tabular}{@{}lccccccccc@{}}
\toprule
\textbf{Methods} & \textbf{R@1 ($\uparrow$)} & \textbf{R@5 ($\uparrow$)} & \textbf{R@10 ($\uparrow$)} & \textbf{MdR ($\downarrow$)} & \textbf{MnR ($\downarrow$)} & \textbf{AP@1 ($\uparrow$)} & \textbf{AP@5 ($\uparrow$)} & \textbf{AP@10 ($\uparrow$)} & \textbf{mAP ($\uparrow$)} \\
\midrule
DGL~\cite{dgl}     & 14.5\tiny{$\pm$1.18}& 38.7\tiny{$\pm$2.63}& 50.3\tiny{$\pm$2.85}& 10.0\tiny{$\pm$1.96}& 26.6\tiny{$\pm$1.61}&14.5\tiny{$\pm$1.18}& 23.1\tiny{$\pm$1.49}& 24.6\tiny{$\pm$1.47}& 26.2\tiny{$\pm$1.38}\\
EERC~\cite{eerc}          & 15.7\tiny{$\pm$1.50}& 30.9\tiny{$\pm$2.17}& 35.1\tiny{$\pm$2.37}& 34.0\tiny{$\pm$4.16}& 48.4\tiny{$\pm$1.86}&15.7\tiny{$\pm$1.50}& 21.8\tiny{$\pm$1.58}& 22.3\tiny{$\pm$1.61}& 23.8\tiny{$\pm$1.55}\\
DiffusionRet~\cite{diffusionret} & 3.8\tiny{$\pm$0.55}& 7.2\tiny{$\pm$1.21}& 11.0\tiny{$\pm$1.62}& 65.0\tiny{$\pm$4.10}& 64.4\tiny{$\pm$2.47}&3.8\tiny{$\pm$0.55}& 4.8\tiny{$\pm$0.63}& 5.3\tiny{$\pm$0.66}& 7.2\tiny{$\pm$0.64}\\
TeachCLIP~\cite{teacherclip}     & 11.0\tiny{$\pm$2.26}& 41.8\tiny{$\pm$2.62}& 55.9\tiny{$\pm$2.27}& 8.0\tiny{$\pm$0.98}& 16.0\tiny{$\pm$0.80}&11.0\tiny{$\pm$2.26}& 22.0\tiny{$\pm$2.16}& 23.8\tiny{$\pm$2.05}& 25.8\tiny{$\pm$1.99}\\
InternVideo2~\cite{internvideo2}  & 17.4\tiny{$\pm$1.47}& 43.2\tiny{$\pm$1.83}& 61.1\tiny{$\pm$1.31}& 7.0\tiny{$\pm$0.46}& 15.9\tiny{$\pm$1.11}&17.4\tiny{$\pm$1.47}& 26.1\tiny{$\pm$1.63}& 28.5\tiny{$\pm$1.55}& 30.2\tiny{$\pm$1.52}\\
Qwen2.5-VL~\cite{qwenvl}      & 26.6\tiny{$\pm$2.12}  & 50.1\tiny{$\pm$2.31} & 63.8\tiny{$\pm$2.06}  & 5.0\tiny{$\pm$0.90}   & \underline{15.3}\tiny{$\pm$1.02}   &26.6\tiny{$\pm$2.12} & 35.6\tiny{$\pm$2.09} & 37.5\tiny{$\pm$1.99} & 39.0\tiny{$\pm$1.95} \\
Gemini2.5~\cite{gemini}        & 7.4\tiny{$\pm$1.40}   & 15.0\tiny{$\pm$1.75}   & 26.4\tiny{$\pm$2.45}   & 25.0\tiny{$\pm$3.86} & 45.2\tiny{$\pm$2.15}   & 7.4\tiny{$\pm$1.40}   & 9.9\tiny{$\pm$1.42}  & 11.4\tiny{$\pm$1.45}  & 13.7\tiny{$\pm$1.35} \\
CLIP4Clip~\cite{luo2022clip4clip}     & \underline{29.3}\tiny{$\pm$2.04}& \underline{56.2}\tiny{$\pm$2.03}& \underline{66.0}\tiny{$\pm$2.06}& \underline{4.0}\tiny{$\pm$0.30}& 15.8\tiny{$\pm$1.29}&\underline{29.3}\tiny{$\pm$2.04}& \underline{38.9}\tiny{$\pm$1.79}& \underline{40.2}\tiny{$\pm$1.80}& \underline{41.4}\tiny{$\pm$1.74}\\
\hline
Ours & \textbf{81.2}\tiny{$\pm$1.81}& \textbf{93.5}\tiny{$\pm$1.24}& \textbf{95.5}\tiny{$\pm$0.97}& \textbf{1.0}\tiny{$\pm$0.01}& \textbf{2.3}\tiny{$\pm$0.30}& \textbf{81.2}\tiny{$\pm$1.81}& \textbf{86.2}\tiny{$\pm$1.51}& \textbf{86.5}\tiny{$\pm$1.51}&\textbf{86.8}\tiny{$\pm$1.46}\\
\bottomrule
\end{tabular}%
}
\end{table*}

\paragraph{Coarse-Grained Retrieval via Compositional Alignment}
Applying LLM to reason on all $N$ videos would be computationally costly, which therefore requires efficient filtering to identify promising candidates.
This filtering presents two challenges for learning cross-modal alignment: (1) implicit queries often reference specific objects or relationships rather than entire videos, making global video-level embeddings too coarse; (2) the semantic gap between implicit query and explicit digital twin representation complicates direct alignment via embeddings.
Rather than encoding the entire digital twin representation $\mathcal{D}_V$ as a single embedding, we therefore decompose it into object-level and relationship-level components, producing a set of embeddings $\mathbf{E}_V = \{\mathbf{e}_{V}^{\text{obj}_1}, ..., \mathbf{e}_{V}^{\text{obj}_{N}}, \mathbf{e}_{V}^{\text{rel}_1}, ..., \mathbf{e}_{V}^{\text{rel}_{M}}\}$,
where $N_V$ denotes the total number of object instances across all frames in video $V$, and $M_V$ denotes the number of pairwise relationships between objects.
Here, each embedding captures a specific aspect of the digital twin representation that can be independently matched against query conditions.
Object-level embeddings $\mathbf{e}_{V}^{\text{obj}_i} = f_{\text{twin}}^{\text{obj}}((i, c_i, \mathbf{a}_i, \mathbf{s}_i))$ encode individual instances with their semantic categories, attributes, and spatial properties.
Relationship-level embeddings $\mathbf{e}_{V}^{\text{rel}_j} = f_{\text{twin}}^{\text{rel}}((n_i, r_{ij}, n_j))$ capture relationships between a pair of objects such as spatial relations.
For the query side, we employ an LLM to decompose the implicit query $q$ into atomic sub-queries $\{q_1, q_2, ..., q_L\}$.
Each sub-query represents a verifiable explicit condition such as object attributes (\textit{e}.\textit{g}., ``an animal''), spatial relationships (\textit{e}.\textit{g}., ``behind the table''), or temporal patterns (\textit{e}.\textit{g}., ``initially approaching'').
We encode each sub-query through the query encoder $f_{\text{query}}: q_l \rightarrow \mathbf{e}_{q_l}$, producing normalized embeddings $\mathbf{E}_q = \{\mathbf{e}_{q_1}, ..., \mathbf{e}_{q_L}\}$.

During training of the paired encoders, we optimize both the object/relationship encoders and the query encoder through a compositional contrastive loss that enforces alignment between sub-queries and their corresponding digital twin representation components $\mathcal{L} = \sum_{l=1}^L \mathcal{L}_{\text{NCE}}(\mathbf{e}_{q_l}, \mathbf{E}_V^+, \mathbf{E}_V^-)$,
where $\mathbf{E}_V^+$ contains embeddings of digital twin representation components satisfying sub-query $q_l$ and $\mathbf{E}_V^-$ contains negative components from other videos.
During inference, we first decompose the implicit query $q$ into sub-queries $\{q_1, ..., q_L\}$ using the LLM, then compute similarities between each sub-query embedding and all digital twin representation component embeddings in the database.
For each video $V_i$, we aggregate results across sub-queries to obtain a compositional similarity score:
\begin{equation}
s_i = \text{Agg}(\{\max_{\mathbf{e} \in \mathbf{E}_{V_i}} \mathbf{e}_{q_l}^\top \mathbf{e}\}_{l=1}^L),
\end{equation}
where the aggregation function combines maximum similarities across sub-queries (\textit{e}.\textit{g}., weighted average or minimum for conjunctive reasoning).
We retrieve the top-$K$ candidates $\mathcal{C} = \{(V_{i_1}, \mathcal{D}_{V_{i_1}}), ..., (V_{i_K}, \mathcal{D}_{V_{i_K}})\}$ with highest compositional scores.

\paragraph{Fine-Grained Retrieval via Reasoning and Reranking}
The top-$K$ candidates $\mathcal{C}$ obtained from compositional alignment require further verification through reasoning.
While embedding-based matching identifies videos containing relevant components, it cannot verify whether these components satisfy the complete set of query conditions or resolve complex logical relationships between sub-queries. 
Consequently, for each candidate video $V_i$ with its digital twin representation $\mathcal{D}{V_i}$, we prompt LLM to reason through a chain of thought.
Specifically, the LLM first identify if there are information gaps in the initial digital twin representation to verify certain query conditions in the sub-query.
When such gaps are detected, we adapt the just-in-time refinement approach~\cite{jit}, where the LLM generates execution plans that specify additional specialist models to invoke (\textit{e}.\textit{g}., vision-language models for detailed attribute descriptions, action recognition models for temporal patterns).
We then execute these requests to augment $\mathcal{D}_{V}$ universally, producing an enriched representation $\mathcal{D}_{V}'$ before the LLM resumes the reasoning.
Afterwards, the LLM produces a structured output for each candidate video containing three components: a relevance score $r_i \in [0, 1]$ indicating how well the video satisfies the query; a reasoning trace explaining which objects and relationships support this judgment; and a set of object identifiers $\mathcal{O}_i = \{o_1, o_2, ..., o_P\}$ representing instances that fulfill the query.
We filter candidates with relevance scores below a threshold $\tau$, as low scores indicate videos that do not genuinely satisfy the query despite passing the coarse-grained retrieval stage. 
The remaining candidates are reranked according to their relevance scores to produce the final ranked list.
Consequently, we extracted mask paths $\{m_{o_1}, m_{o_2}, ..., m_{o_P}\}$ corresponding to the identified object identifiers from the digital twin representation to denote the specific regions in the retrieved videos.

\begin{table*}[t!]
\centering
\caption{Performance evaluation on \textit{ReasonT2VBench-1000} with 865 distractor videos. This extended benchmark assesses semantic discrimination capability in large-scale retrieval scenarios. Our approach maintains robust performance despite visual distractors, while baseline methods show degradation. \textbf{Bold} indicates best results and \underline{underline} indicates second-best results.}
\label{tab:extended_model_comparison}
\resizebox{\linewidth}{!}{%
\begin{tabular}{@{}lccccccccccccc@{}}
\toprule
\textbf{Methods} & \textbf{R@1 ($\uparrow$)} & \textbf{R@5 ($\uparrow$)} & \textbf{R@10 ($\uparrow$)} & \textbf{R@50 ($\uparrow$)} & \textbf{R@100 ($\uparrow$)} & \textbf{MdR ($\downarrow$)} & \textbf{MnR ($\downarrow$)} & \textbf{AP@1 ($\uparrow$)} &  \textbf{AP@5 ($\uparrow$)} & \textbf{AP@10 ($\uparrow$)} & \textbf{AP@50 ($\uparrow$)} & \textbf{AP@100 ($\uparrow$)} & \textbf{mAP ($\uparrow$)} \\
\midrule
DGL~\cite{dgl}    & 8.1\tiny{$\pm$0.97}   & 17.4\tiny{$\pm$1.60}  & 23.9\tiny{$\pm$1.89}  & 49.9\tiny{$\pm$1.58} & 66.9\tiny{$\pm$1.75} & 51.0\tiny{$\pm$3.00}  & 117.9\tiny{$\pm$4.96}  &8.1\tiny{$\pm$0.97}   
& 11.2\tiny{$\pm$1.10}& 11.9\tiny{$\pm$1.09}& 12.6\tiny{$\pm$1.05}& 12.8\tiny{$\pm$1.05}& 13.7\tiny{$\pm$1.05}\\
EERC~\cite{eerc}        & 8.1\tiny{$\pm$0.63}   & 22.8\tiny{$\pm$1.17}   & 25.8\tiny{$\pm$1.23}   & 40.0\tiny{$\pm$1.21}   & 52.9\tiny{$\pm$1.25}   & 223.0\tiny{$\pm$12.84} & 283.2\tiny{$\pm$8.29}  &8.1\tiny{$\pm$0.63}   
& 13.7\tiny{$\pm$0.75} & 14.1\tiny{$\pm$0.75} & 14.3\tiny{$\pm$0.74}& 14.3\tiny{$\pm$0.74}& 14.5\tiny{$\pm$0.74}\\
DiffusionRet~\cite{diffusionret}  & 0.7\tiny{$\pm$0.36}& 3.8\tiny{$\pm$1.02}& 7.6\tiny{$\pm$1.16}& 34.0\tiny{$\pm$1.67}& \underline{72.9}\tiny{$\pm$1.27}& 75.0\tiny{$\pm$1.85}& \underline{70.5}\tiny{$\pm$1.61}&0.7\tiny{$\pm$0.36}   
& 1.7\tiny{$\pm$0.48}& 2.1\tiny{$\pm$0.50}& 3.2\tiny{$\pm$0.49}& 3.8\tiny{$\pm$0.48}& 4.0\tiny{$\pm$0.48}\\
TeachCLIP~\cite{teacherclip}     & 3.1\tiny{$\pm$0.79}   & 6.7\tiny{$\pm$0.99}   & 9.6\tiny{$\pm$1.39}   & 23.7\tiny{$\pm$2.80}   & 34.2\tiny{$\pm$3.15}   & 267.0\tiny{$\pm$40.22} & 332.1\tiny{$\pm$15.07}  &3.1\tiny{$\pm$0.79}   
& 4.4\tiny{$\pm$0.83}& 4.7\tiny{$\pm$0.84}& 5.3\tiny{$\pm$0.85}& 5.5\tiny{$\pm$0.85}& 5.6\tiny{$\pm$0.85}\\
InternVideo2~\cite{internvideo2}  & 10.3\tiny{$\pm$1.08}& 25.3\tiny{$\pm$2.65}& 30.4\tiny{$\pm$2.98}& 56.6\tiny{$\pm$2.66}& 69.1\tiny{$\pm$1.72}& 37.0\tiny{$\pm$5.89}& 117.2\tiny{$\pm$5.78}&10.3\tiny{$\pm$1.08}& 15.9\tiny{$\pm$1.53}& 16.6\tiny{$\pm$1.54}& 17.8\tiny{$\pm$1.55}& 18.0\tiny{$\pm$1.54}& 18.1\tiny{$\pm$1.54}\\
Qwen2.5-VL~\cite{qwenvl}     & 8.5\tiny{$\pm$1.01}   & 23.9\tiny{$\pm$1.83}  & 33.6\tiny{$\pm$1.97}  & \underline{58.6}\tiny{$\pm$2.23}  & 69.6\tiny{$\pm$1.85}  & 24.0\tiny{$\pm$1.32}  & 106.3\tiny{$\pm$2.60}  &8.5\tiny{$\pm$1.01}   
& 14.3\tiny{$\pm$1.45}& 15.4\tiny{$\pm$1.47}& 16.6\tiny{$\pm$1.46}& 16.9\tiny{$\pm$1.46}& 17.0\tiny{$\pm$1.46} \\
Gemini2.5~\cite{gemini}        & 7.4\tiny{$\pm$0.01}   & 15.0\tiny{$\pm$3.18}   & 26.4\tiny{$\pm$4.02}   & 25.0\tiny{$\pm$6.95}  & 19.2\tiny{$\pm$8.32}  & 342.0\tiny{$\pm$0.01} & 397.9\tiny{$\pm$58.64}  &1.1\tiny{$\pm$0.01}   & 1.9\tiny{$\pm$0.96}& 2.0\tiny{$\pm$0.98}& 2.1\tiny{$\pm$0.99}& 2.1\tiny{$\pm$1.01}& 2.1\tiny{$\pm$1.01}\\
CLIP4Clip~\cite{luo2022clip4clip}     & \underline{24.9}\tiny{$\pm$1.80} & \underline{41.0}\tiny{$\pm$1.58}  & \underline{45.5}\tiny{$\pm$1.41}  & 55.9\tiny{$\pm$1.73}  & 60.5\tiny{$\pm$1.58}  & \underline{20.0}\tiny{$\pm$6.83}  & 212.6\tiny{$\pm$9.40}  &\underline{24.9}\tiny{$\pm$1.80} 
& \underline{31.3}\tiny{$\pm$1.68}& \underline{31.9}\tiny{$\pm$1.64}& \underline{32.3}\tiny{$\pm$1.63}& \underline{32.4}\tiny{$\pm$1.62}& \underline{32.5}\tiny{$\pm$1.61}\\
\hline
Ours & \textbf{81.7}\tiny{$\pm$2.01}& \textbf{92.2}\tiny{$\pm$1.24}& \textbf{94.0}\tiny{$\pm$1.20}& \textbf{97.5}\tiny{$\pm$0.53}& \textbf{98.9}\tiny{$\pm$0.38}& \textbf{1.0}\tiny{$\pm$0.01}& \textbf{7.2}\tiny{$\pm$1.64}&\textbf{81.7}\tiny{$\pm$2.01}& \textbf{85.8}\tiny{$\pm$1.52}& \textbf{86.1}\tiny{$\pm$1.51}& \textbf{86.2}\tiny{$\pm$1.49}& \textbf{86.2}\tiny{$\pm$1.49}& \textbf{85.2}\tiny{$\pm$0.72}\\
\bottomrule
\end{tabular}%
}
\end{table*}

\paragraph{Benchmark Dataset}
To evaluate reasoning text-to-video retrieval, we construct a benchmark with two configurations, namely \textit{ReasonT2VBench-135} and \textit{ReasonT2VBench-1000}
.
The base configuration, \textit{ReasonT2VBench-135}, contains 135 videos sourced from the SA-V test set~\cite{sam2}, recorded at 24 FPS with a mean resolution of 1262$\times$936 pixels.
We leverage the instance-level ground-truth masks from SA-V~\cite{sam2} to guide the manual creation of 447 implicit text queries that necessitate reasoning for retrieval.
Consequently, each video in \textit{ReasonT2VBench-135} is paired with one or more implicit queries.
\textit{ReasonT2VBench-1000} augments the base configuration with 865 additional distractor videos from Youtube that exhibit visual similarities to the annotated videos but have no associated text queries.
These distractors test the t2v retrieval method's ability to distinguish between visually similar content that differs semantically.
To guarantee one-to-one video-query correspondence, we also implement a two-stage validation pipeline after the benchmark creation, via initial screening through Twelve Labs' Marengo 2.7
, which is a video retrieval model, followed by manual verification by human annotators.
\begin{table*}[t!]
\centering
\caption{Ablation study of the components in our reasoning text-to-video retrieval framework. 
QD: query decomposition into sub-queries, CA: compositional alignment for coarse-grained retrieval, LR: LLM-based reasoning for fine-grained retrieval, JiT: Just-in-Time refinement. 
}
\label{tab:ablation}
\resizebox{\textwidth}{!}{%
\begin{tabular}{@{}cccc|ccccccc|ccccccc@{}}
\toprule
\multirow{2}{*}{QD} & \multirow{2}{*}{CA} & \multirow{2}{*}{LR} & \multirow{2}{*}{JiT} & \multicolumn{7}{c|}{\textbf{ReasonT2VBench-135}} & \multicolumn{7}{c}{\textbf{ReasonT2VBench-1000}} \\
\cmidrule(lr){5-11} \cmidrule(lr){12-18}
& & & & R@1 & R@5 & R@10 & MnR & AP@1 & AP@5 & mAP & R@1 & R@5 & R@10 & MnR & AP@1 & AP@5 & mAP \\
\hline
\texttimes & \checkmark & \checkmark & \checkmark &  75.8\tiny{$\pm$1.25}&  90.6\tiny{$\pm$1.07}&  92.8\tiny{$\pm$0.67}&  3.2\tiny{$\pm$0.39}&  75.8\tiny{$\pm$1.25}&  82.1\tiny{$\pm$0.9}&  82.8\tiny{$\pm$0.88}& 70.5\tiny{$\pm$1.97}& 86.4\tiny{$\pm$1.28}& 90.4\tiny{$\pm$1.34}& 11.5\tiny{$\pm$2.10}& 70.5\tiny{$\pm$1.97}& 77.1\tiny{$\pm$1.59}& 78.0\tiny{$\pm$1.54}\\
\checkmark & \texttimes & \checkmark & \checkmark & 78.9\tiny{$\pm$1.50}&  88.8\tiny{$\pm$1.03}&  89.5\tiny{$\pm$0.85}&  3.6\tiny{$\pm$0.24}&  78.9\tiny{$\pm$1.50}&  82.5\tiny{$\pm$1.25}&  83.1\tiny{$\pm$1.20}& 74.7\tiny{$\pm$2.97}& 83.2\tiny{$\pm$2.61}& 84.1\tiny{$\pm$2.41}& 16.5\tiny{$\pm$3.19}& 74.7\tiny{$\pm$2.97}& 78.2\tiny{$\pm$2.84}& 78.6\tiny{$\pm$2.77}\\
\checkmark & \checkmark & \texttimes & \checkmark &  71.6\tiny{$\pm$2.15}&  90.2\tiny{$\pm$1.52}&  95.5\tiny{$\pm$1.09}&  2.7\tiny{$\pm$0.25}&  71.6\tiny{$\pm$2.15}&  78.3\tiny{$\pm$1.87}&  79.4\tiny{$\pm$1.75}&  64.7\tiny{$\pm$2.02}&  86.1\tiny{$\pm$1.34}&  91.3\tiny{$\pm$1.42}&  7.7\tiny{$\pm$1.84}&  64.7\tiny{$\pm$2.02}&  73.1\tiny{$\pm$1.62}&  74.2\tiny{$\pm$1.64}\\
\checkmark & \checkmark & \checkmark & \texttimes & 75.2\tiny{$\pm$2.28}& 88.1\tiny{$\pm$1.23}& 91.3\tiny{$\pm$1.22}&  3.9\tiny{$\pm$0.49}&  75.2\tiny{$\pm$2.28}& 80.3\tiny{$\pm$1.66}&  81.2\tiny{$\pm$1.57}& 72.7\tiny{$\pm$1.70}& 84.8\tiny{$\pm$1.45}& 87.0\tiny{$\pm$1.27}& 17.6\tiny{$\pm$2.27}&  72.7\tiny{$\pm$1.70}& 77.6\tiny{$\pm$1.40}& 78.1\tiny{$\pm$1.35}\\
\hline
\checkmark & \checkmark & \checkmark & \checkmark &  81.2\tiny{$\pm$1.81}&  93.5\tiny{$\pm$0.92}&  95.5\tiny{$\pm$0.97}&  2.3\tiny{$\pm$0.30}&  81.2\tiny{$\pm$1.81}&  86.2\tiny{$\pm$1.51}&  86.8\tiny{$\pm$1.46}& 81.7\tiny{$\pm$2.01}& 92.2\tiny{$\pm$1.24}& 94.0\tiny{$\pm$1.20}& 7.2\tiny{$\pm$1.64}& 81.7\tiny{$\pm$2.01}& 85.8\tiny{$\pm$1.52}& 86.2\tiny{$\pm$1.49}\\
\bottomrule
\end{tabular}%
}
\end{table*}

\section{Experiments}

\paragraph{Implementation Details}
We implement all experiments in PyTorch 3.12 and conduct experiments on 8 NVIDIA GeForce RTX 4090 GPUs with 24GB memory each. 
For query decomposition and reasoning, we employ GPT-5 as the LLM backbone, while the paired encoders for compositional alignment are based on the CLIP architecture with two separate BERTs~\cite{clip} trained using the \textit{RVTBench} reasoning segmentation subset~\cite{rvtbench}.
During coarse-grained retrieval, we retrieve the top-$K=10$ candidates, and apply a relevance threshold of $\tau=0.5$ for fine-grained filtering.

\paragraph{Evaluation Metrics}
Our evaluation considers two aspects: (1) retrieval quality, which assesses whether correct videos are identified and ranked appropriately, and (2) grounding quality, which measures whether target objects are accurately localized within retrieved videos.
Following standard practices in text-to-video retrieval, we adopt ranking-based metrics to assess model retrieval performance~\cite{rodriguez2022fighting}. 
For ranking quality assessment, we report Recall at rank K (R@K) for K values of 1, 5, 10, 50, and 100, which measures the proportion of queries where at least one relevant video appears within the top K retrieved results. 
Additionally, we compute the median rank (MdR) and mean rank (MnR), representing the median and average positions of the first relevant video in all queries, respectively, where lower values indicate better performance~\cite{wang2022multi}. 
To capture ranking quality beyond recall-based metrics, we employ Average Precision at K (AP@K) for K values of 1, 5, 10, 50, and 100, which calculates precision scores at positions where relevant items appear and averages them across queries. 
The overall mean Average Precision (mAP) provides a summary of a single value by averaging AP scores across all K values. 
For grounding accuracy, we adopt region similarity ($\mathcal{J}$) and contour accuracy ($\mathcal{F}$) following the reasoning segmentation~\cite{survey}. 
The region similarity $\mathcal{J}$ computes the intersection-over-union (IoU) between the predicted and ground-truth masks, quantifying the precision of pixel-level segmentation in a scale-invariant manner. 
The contour accuracy $\mathcal{F}$ evaluates the boundary precision and recall through bipartite matching between the predicted and ground-truth object contours.

\paragraph{Compared Methods}
We compare our approach with various text-to-video retrieval methods. 
CLIP4Clip~\cite{luo2022clip4clip} adapts pre-trained image-text models to video domains through frame-level feature aggregation, computing relevance via cosine similarity between query and video embeddings in a shared space.
DGL~\cite{dgl} introduces cross-modal dynamic prompt tuning with global-local video attention.
TeachCLIP~\cite{teacherclip} proposes multi-grained teaching where a CLIP4Clip-based student network learns from computationally expensive teacher models through attentional frame-feature aggregation that imitates frame-text relevance.
EERC~\cite{eerc} focuses on learning coarse-to-fine visual representation with a parameter-free text-gated interaction block for cross-modal optimization.
Video-LLM approaches including InternVideo2~\cite{internvideo2}, Qwen2.5-VL~\cite{qwenvl}, and Gemini2.5~\cite{gemini} represent video content through compressed token sequences to fit within LLM context windows.
DiffusionRet~\cite{diffusionret} employs generative text-video retrieval through diffusion models to compute relevance scores between queries and videos.

\begin{figure*}[!t]
\centering
\includegraphics[width=\linewidth]{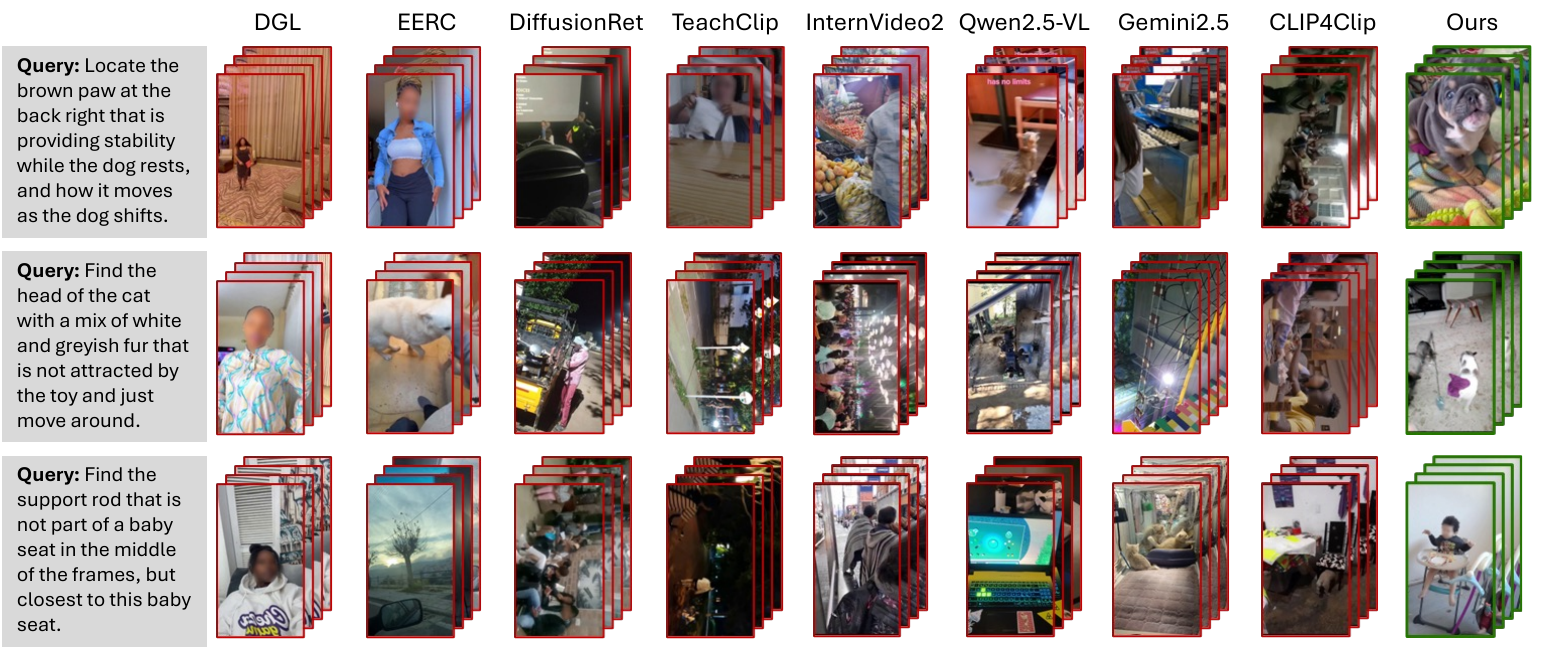}
\caption{Qualitative comparison of reasoning text-to-video retrieval across three implicit queries on \textit{ReasonT2VBench-135}. 
Red borders indicate incorrect or failed retrievals where methods return irrelevant videos, while green borders show correct retrieval and grounding.
Our method correctly identifies relevant videos and grounds the target objects via segmentation masks (in purple).
}
\label{fig:result}
\end{figure*}

\begin{table*}[t!]
\centering
\caption{Performance comparison on three conventional text-to-video retrieval benchmarks.}
\label{tab:conventional_retrieval}
\resizebox{\linewidth}{!}{%
\begin{tabular}{@{}lccccc|ccc|ccc@{}}
\toprule
& \multicolumn{5}{c|}{\textbf{MSR-VTT}~\cite{msrvtt}} & \multicolumn{3}{c|}{\textbf{MSVD}~\cite{msvd}} & \multicolumn{3}{c}{\textbf{VATEX}~\cite{vatex}} \\
\cmidrule(lr){2-6} \cmidrule(lr){7-9} \cmidrule(lr){10-12}
\textbf{Methods} & R@1 ($\uparrow$) & R@5 ($\uparrow$) & R@10 ($\uparrow$) & MdR ($\downarrow$) & MnR ($\downarrow$) & R@1 ($\uparrow$) & R@5 ($\uparrow$) & R@10 ($\uparrow$) & R@1 ($\uparrow$) & R@5 ($\uparrow$) & R@10 ($\uparrow$) \\
\midrule
DGL~\cite{dgl} & 14.5 & 38.7 & 50.3 & 9.0 & 25.3 & 32.6 & 64.8 & 78.1 & 59.8 & 87.2 & 92.1\\
EERC~\cite{eerc} & 15.7 & 30.9 & 35.1 & 15.0 & 42.7 & 27.2 & 55.1 & 65.6 & 43.8 & 76.5 & 85.2 \\
DiffusionRet~\cite{diffusionret} & 22.4 & 48.9 & 60.2 & 4.0 & 48.9 & 40.8 & 69.9 & 78.9 & 50.6 & 82.7 & 90.1 \\
TeachCLIP~\cite{teacherclip} & 11.0 & 41.8 & 55.9 & 6.0 & 18.4 & 30.2 & 60.8 & 72.4 & 63.6 & 91.9 & 96.1 \\
InternVideo2~\cite{internvideo2} & \underline{55.2} & \underline{80.1} & \underline{87.8} & \textbf{1.0} & \underline{4.8} & \underline{62.8} & \underline{86.9} & \underline{93.1} & \underline{75.5} & \underline{95.1} & \underline{97.8} \\
Qwen2.5-VL~\cite{qwenvl} & 26.6 & 50.1 & 63.8 & 3.0 & 12.1 & 42.1 & 71.3 & 81.7 & 61.5 & 87.2 & 93.4 \\
Gemini2.5~\cite{gemini} & 18.7 & 42.3 & 55.2 & 7.0 & 21.8 & 28.9 & 58.4 & 71.2 & 45.7 & 78.1 & 87.3 \\
CLIP4Clip~\cite{luo2022clip4clip} & 43.4 & 70.2 & 80.6 & \underline{2.0} & 9.6 & 47.3 & 76.8 & 85.9 & 46.2 & 79.6 & 88.1 \\
\hline
\textbf{Ours} & \textbf{58.9} & \textbf{83.8} & \textbf{91.2} & \textbf{1.0} & \textbf{4.0} & \textbf{66.5} & \textbf{90.4} & \textbf{96.5} & \textbf{79.2} & \textbf{98.4} & \textbf{99.2} \\
\bottomrule
\end{tabular}%
}
\end{table*}

\paragraph{Evaluation on ReasonT2VBench-135}
Table~\ref{tab:model_comparison} presents performance comparisons on \textit{ReasonT2VBench-135}.
Our method achieves 81.2\% R@1, representing a 51.9\% gain over CLIP4Clip~\cite{luo2022clip4clip}, which achieves 29.3\% as the best-performing baseline. 
The performance improvement of our method over baselines expands at higher ranks, where our approach reaches 93.5\% R@5 and 95.5\% R@10, whereas CLIP4Clip yields 56.2\% and 66.0\% respectively. 
%
%
In average precision metrics, our method obtains 86.8\% mAP against CLIP4Clip's 41.4\%, indicating better ranking quality throughout the retrieved video set.
Other embedding-based methods, including DGL~\cite{dgl} and EERC~\cite{eerc} achieve R@1 scores below 16\%, reflecting their limitations in decomposing implicit queries into structured reasoning steps. 
DiffusionRet~\cite{diffusionret} performs worse at 3.8\% R@1, indicating that the generative retrieval paradigms struggle with reasoning-driven video search tasks. 
Video-LLM approaches, including InternVideo2, Qwen2.5-VL, and Gemini2.5, produce R@1 scores between 7.4\% and 26.6\%, showing how token compression fails to preserve the spatial and object-level detail necessary for grounding implicit queries during the retrieval process.
Fig.~\ref{fig:result} illustrates these retrieval failures visually, where baseline methods frequently return videos that do not satisfy the query conditions despite containing superficially similar visual elements.
The poor performance across all video-LLM variants, despite their strong natural language understanding capabilities, confirms that preserving fine-grained visual structure through digital twin representations is necessary for reasoning retrieval tasks
These findings validate our two-stage framework, where compositional alignment filters candidates efficiently and LLM-based reasoning verifies complex query conditions against structured scene representations.

\paragraph{Evaluation on ReasonT2VBench-1000}
Table~\ref{tab:extended_model_comparison} presents the results on \textit{ReasonT2VBench-1000}.
Our method maintains 81.7\% R@1, demonstrating stability when scaling from 135 to 1000 candidates. 
In contrast, CLIP4Clip drops from 29.3\% to 24.9\% R@1, exposing how embedding-based approaches fail to distinguish semantically relevant videos from visually similar but irrelevant content. 
This degradation reveals that holistic embeddings capture surface-level visual patterns without understanding the semantic relationships required for implicit query matching.
On the other hand, video-LLM methods experience a greater performance drop, where InternVideo2~\cite{internvideo2} falls from 17.4\% to 10.3\% R@1, while Qwen2.5-VL~\cite{qwenvl} decreases from 17.2\% to 8.5\%. 
These declines confirm that token compression removes the object-level information needed to differentiate between videos that appear visually similar, but differ in semantic content. 
The compressed representations lack sufficient granularity to resolve fine-grained distinctions when presented with challenging distractors during the reasoning text-to-video retrieval.
The mean rank metric reveals different scaling behaviors between methods. 
Our method's MnR increases modestly from 2.3 to 7.2, whereas CLIP4Clip's MnR rises dramatically from 15.8 to 212.6, which demonstrates that our LLM-based reasoning stage filters false positives during reranking, preventing visually similar distractors from displacing truly relevant results. 
The consistent performance across both benchmark configurations shows that our method scales effectively to databases resembling real-world video retrieval scenarios where semantic precision must be maintained despite visual ambiguity.

\paragraph{Evaluation on Conventional Text-to-Video Retrieval Benchmarks}
Although our method targets reasoning retrieval with implicit queries, we also evaluate on three conventional text-to-video retrieval benchmarks.
Specifically, we leverage (1) MSR-VTT~\cite{msrvtt}, which contains 10,000 web video clips with 200,000 clip-sentence pairs in 20 categories; (2) MSVD~\cite{msvd} (YouTube2Text), which consists of 1,970 YouTube clips with 80,000 captions; and (3) VATEX~\cite{vatex}, which has 41,250 videos with 825,000 English-Chinese parallel captions covering 600 human activities.
These three benchmarks use explicit queries that directly describe visual content.
Table~\ref{tab:conventional_retrieval} presents the results.
Our method achieves the best performance across all three benchmarks, outperforming the second-best approach InternVideo2 by 3.7\% on MSR-VTT R@1, 3.7\% on MSVD R@1, and 3.7\% on VATEX R@1.
The consistent improvements demonstrate that our digital twin representation preserves fine-grained visual details that benefit retrieval even for explicit queries.

\paragraph{Ablation Study on Retrieval Performance}

Table~\ref{tab:ablation} shows the ablation study on how each component contributes to the overall retrieval performance. 
Specifically, without query decomposition, R@1 falls to 75.8\% on \textit{ReasonT2VBench-135} and 70.5\% on \textit{ReasonT2VBench-1000}, representing drops of 5.4\% and 11.2\% respectively. 
The higher degradation on \textit{ReasonT2VBench-1000} indicates that the structured query representation becomes more important when the model needs to distinguish relevant videos from numerous visually similar distractors. 
%
%
Performance without compositional alignment drops to 78.9\% on \textit{ReasonT2VBench-135} and 74.7\% on \textit{ReasonT2VBench-1000} as these learned embeddings enable efficient candidate filtering before applying computationally expensive reasoning.
The 7.0\% drop on \textit{ReasonT2VBench-1000} confirms that this coarse-grained retrieval becomes necessary as search spaces expand.
%
%
Removing the LLM-based reasoning causes R@1 to fall to 71.6\% on \textit{ReasonT2VBench-135} and 64.7\% on \textit{ReasonT2VBench-1000}, while mean rank increases from 2.3 to 2.7 and from 7.2 to 17.6 respectively. 
%
%
%
Without just-in-time refinement, R@1 reaches 75.2\% on \textit{ReasonT2VBench-135} and 72.7\% on \textit{ReasonT2VBench-1000}. 
%
%
Consequently, our ablation reveals that all four components deliver substantial gains to the overall performance of our method.

\begin{table}[ht!]
\centering
\caption{Ablation study on object-level grounding quality. 
}
\label{tab:ablation_grounding}
\resizebox{\linewidth}{!}{%
\begin{tabular}{@{}cccccccc@{}}
\toprule
\multirow{2}{*}{QD} & \multirow{2}{*}{CA} & \multirow{2}{*}{LR} & \multirow{2}{*}{JiT} & \multicolumn{2}{c}{\textbf{ReasonT2VBench-135}} & \multicolumn{2}{c}{\textbf{ReasonT2VBench-1000}} \\
\cmidrule(lr){5-6} \cmidrule(lr){7-8}
& & & & $\mathcal{J}$ & $\mathcal{F}$ & $\mathcal{J}$ & $\mathcal{F}$ \\
\midrule
\texttimes & \checkmark & \checkmark & \checkmark & 52.4 & 54.0 & 48.7 & 50.3 \\
\checkmark & \texttimes & \checkmark & \checkmark & 52.2 & 54.1 & 50.3 & 52.3 \\
\checkmark & \checkmark & \texttimes & \checkmark & 47.4 & 49.0 & 42.6 & 44.0 \\
\checkmark & \checkmark & \checkmark & \texttimes & 51.0 & 52.3 & 50.0 & 51.3 \\
\midrule
\checkmark & \checkmark & \checkmark & \checkmark & \textbf{54.4} & \textbf{56.2} & \textbf{55.0} & \textbf{56.9} \\
\bottomrule
\end{tabular}%
}
\end{table}

\paragraph{Ablation Study on Grounding Quality}

Unlike traditional text-to-video retrieval methods that produce only video-level rankings, our approach generates instance-level segmentation masks identifying which objects within retrieved videos satisfy the query conditions. 
However, all baseline methods lack this object-level grounding capability.
Table~\ref{tab:ablation_grounding} therefore examines how each component of our framework contributes to grounding performance.
LLM-based reasoning has the greatest impact on grounding quality, with its removal causing region similarity to drop by 7.0 and 12.4\% on the two benchmark configurations. 
The complete framework achieves 54.4\% region similarity and 56.2\% contour accuracy on \textit{ReasonT2VBench-135}. 
\section{Conclusion}

We introduce reasoning text-to-video retrieval, a task formulation that extends traditional video retrieval to handle implicit queries demanding reasoning while simultaneously providing object-level grounding masks. 
We propose to decompose videos into structured scene descriptions (\textit{i}.\textit{e}., digital twin representation) through specialist vision models, enabling LLMs to reason directly over textual representations without visual token compression.
Correspondingly, we propose a two-stage method for reasoning text-to-video retrieval task, where compositional alignment first filters candidates efficiently and then is filtered by LLM.
Future work could extend this approach to cross-modal retrieval scenarios that involve audio patterns, motion dynamics. 
%

\newpage

{
    \small
    \bibliographystyle{ieeenat_fullname}
    \bibliography{main}
}


\end{document}